\documentclass[conference]{IEEEtran}
\IEEEoverridecommandlockouts

\usepackage{amsmath,amssymb,amsfonts}
\usepackage{graphicx}
\usepackage{textcomp}
\usepackage{physics}
\usepackage{enumerate}
\usepackage{algorithm}
\usepackage{graphicx}
\usepackage{algpseudocode}
\usepackage{color, colortbl, soul}
\usepackage{xcolor}
\usepackage{booktabs}
\usepackage{multicol}
\usepackage{multirow}
\usepackage{subfig}
\usepackage{nccmath}
\usepackage[numbers,sort]{natbib}
\usepackage{caption}
\definecolor{LightCyan}{rgb}{0.78, 1, 1}

\usepackage{color}
\usepackage{xcolor}
\newcommand{\minisection}[1]{\vspace{0.025in} \noindent {\bf #1}\ }

\DeclareMathOperator{\argmin}{arg\,min}

\def\BibTeX{{\rm B\kern-.05em{\sc i\kern-.025em b}\kern-.08em
    T\kern-.1667em\lower.7ex\hbox{E}\kern-.125emX}}
\begin{document}

\title{Generating Personas for Games with Multimodal Adversarial Imitation Learning}


\author{
William Ahlberg,
Alessandro Sestini,
Konrad Tollmar,
Linus Gisslén
\\
\textit{SEED - Electronic Arts (EA)}
\\
\{wahlberg, asestini, ktollmar, lgisslen\}@ea.com
}


\IEEEoverridecommandlockouts

\IEEEpubid{\makebox[\columnwidth]{979-8-3503-2277-4/23/\$31.00~\copyright2023 IEEE \hfill} 
\hspace{\columnsep}\makebox[\columnwidth]{ }}

\maketitle


\IEEEpubidadjcol

\begin{abstract}
Reinforcement learning has been widely successful in producing agents capable of playing games at a human level. However, this requires complex reward engineering, and the agent's resulting policy is often unpredictable. Going beyond reinforcement learning is necessary to model a wide range of human playstyles, which can be difficult to represent with a reward function. This paper presents a novel imitation learning approach to generate multiple persona policies for playtesting. Multimodal Generative Adversarial Imitation Learning (MultiGAIL) uses an auxiliary input parameter to learn distinct personas using a single-agent model. MultiGAIL is based on generative adversarial imitation learning and uses multiple discriminators as reward models, inferring the environment reward by comparing the agent and distinct expert policies. The reward from each discriminator is weighted according to the auxiliary input. Our experimental analysis demonstrates the effectiveness of our technique in two environments with continuous and discrete action spaces.
\end{abstract}


\section{Introduction}
Playtesting is an essential method of quality control. The overall player experience can significantly suffer due to the presence of gameplay issues, such as bugs and glitches, but also because of poor game design choices. Testing makes it possible to iterate game development, and discover and fix problems before launch. Playtests are mainly performed manually by humans who are tasked with playing games, or parts of games, and finding gameplay issues, as well as evaluating if the game is enjoyable and sufficiently challenging. Manual playtesting is a powerful tool, but it is both expensive and time-consuming, especially for large and intricate games. As a result, manual playtesting on its own and when not supported with automated solutions can become expensive, time consuming and increasingly infeasible for large AAA video games. Game designers have to anticipate how differences in player decision making can affect players' experiences. It is of great interest for game designers to be able to model player decision making not only taking into consideration players' skill at a game, but also their specific playstyle, also called \textit{persona}. By incorporating different personas in playtesting, game designers are better suited to foresee how players will interact with the game, and make design choices to, for example, improve game balance.

Machine Learning (ML) presents an opportunity for large video games to be tested with automatic solutions, reducing costs and allowing testing to grow at scale. ML can also be used to model decision making, incorporating an automatic method of validating game design choices for different personas. Self-learning Reinforcement Learning (RL) agents have been successful in creating agents able to play complex games at a human, or even beyond human, level of performance \cite{mnih_playing_2013, vinyals_grandmaster_2019, openai_dota_2019}. At the same time, automated playtesting techniques have been proposed to reduce the reliance on manual validation of large games. Several approaches from the literature are based on model-based automated playtesting~\cite{playfulness, scenariobased}. Moreover, recent research has proposed to train RL agents to effectively cover game states and validate game mechanics \cite{sestini_automated_2022, gordillo_improving_2021}.

Most previously cited works aim to produce an agent which always has the same behavioral style, but cannot generate new personas. One of the main reasons is the difficulty in accurately defining a reward function encompassing qualitative characteristics like a persona. It often requires extensive reward engineering by those with in-depth domain knowledge~\cite{inefficiency, towardstest, microsoftblog}. Imitation Learning (IL) has properties we believe could benefit ML-based playtesting. IL learns from demonstrations rather than experience making it more sample efficient, and can thus circumvent the need for reward engineering. Demonstrations can be done by those who have knowledge of the game, rather than just those with a ML background, making automatic playtesting based on IL accessible to game creators \cite{sestini_towards_nodate}.

We present Multimodal GAIL (MultiGAIL), a novel imitation learning algorithm able to create autonomous agents exhibiting different personas without the use of manual reward engineering. With a single model, MultiGAIL can mimic the policy of a multimodal set of different persona demonstrations, and can switch and blend playstyles by conditioning the policy with an auxiliary input parameter. Game creators can easily and intuitively configure agents by sampling the behavior space induced by the demonstration data at inference. Evaluation of our novel method is done with quantitative and qualitative experimental validation in two distinct environments. The evaluation shows how our approach is effective in both discrete and continuous action spaces. Although MultiGAIL has been developed for training agents suitable for game testing, we believe potential applications extend beyond this purpose. This approach could be useful to anyone looking to easily create distinct behavioral styles without the need for extensive reward engineering for a single RL model.

\section{Related Work}
\label{sec:related}
Recent studies have explored how AI techniques can be used to perform automated playtesting. In this section we present relevant research in relation to our work.

\subsection{Automated Playtesting.}%
Previous works in automated playtesting have mostly focused on classical AI techniques using heuristics and random search algorithms~\cite{playfulness, scenariobased}. Monte-Carlo Tree Search (MCTS) has been shown to be able to master games such as Chess, Shogi and Go \cite{silver_mastering_2017}, but has also been applied to mimic human playstyles in Match-3 games to analyze the performance range of human players \cite{mugrai_automated_2019}. However, complex 3D-environments pose a challenge for these methods due to the high-dimensional state space. The recent success of deep neural networks and Reinforcement Learning (RL) has prompted extensive studies in their potential for automated playtesting. Supervised deep learning has been used to simulate human-like playtesting to predict level difficulty \cite{gudmundsson_human-like_2018}; however, the method is limited by the need for player data at a scale not plausible for new games in production. At the same time RL has been proposed to augment the game testing framework, introducing self-learning to testing to find game exploits and bugs \cite{bergdahl_augmenting_2020}. An implicit reward function, such as curiosity, can also help state coverage of the game space \cite{gordillo_improving_2021}. For commercially available games, the Wuji framework successfully found gameplay issues using RL and evolutionary multi-objective optimization \cite{zheng_wuji_2019}. However, RL is known to require extensive computation to explore its state-action space, while also leaving little room for control over the final policy making it difficult to integrate into any game designers workflow \cite{sestini_towards_nodate}.

\subsection{Personas.}%
Recent works have proposed to combine the game-metrical framework and machine learning to imbue self-learning agents with distinct personas. RL and MCTS have been used to condition agent policies to follow the characteristics of personas, defined with game-metrics. These methods were implemented in 2D discrete environments by engineering their fitness or reward function to condition the policies \cite{holmgard_automated_2018, microsoftblog}. However, leveraging standard RL by defining and shaping a complex reward function like the work by \citet{de_woillemont_configurable_2021}, can be impractical due to the difficulty of a human to define a reward function that captures qualitative behaviors~\cite{amodei_concrete_2016}. Therefore, we decide to employ imitation learning as an alternative.

\subsection{Imitation Learning.}%
Imitation Learning (IL) aims to teach agents to mimic skills and behaviors of an expert in a given task from recorded demonstrations. The work by~\citet{sestini_towards_nodate} indicates that it is a promising avenue for game testing agent. 
Behavioral cloning (BC) is a standard approach that frames IL as a supervised learning problem~\cite{bc1}. However, BC methods suffer from distributional shift when faced with previously unseen trajectories. The DAgger algorithm~\cite{ross_reduction_2011} tries to mitigate this by allowing the agent to query an expert to correct for behaviors not initially taken into account in the demonstration, updating the policy accordingly. Generative Adversarial Imitation Learning (GAIL) \cite{ho_generative_2016} is a technique that combines the notion of adversarial training from Generative Adversarial Networks (GANs) \cite{goodfellow_generative_2014} and imitation learning, a connection formalized by \citet{finn_connection_2016}. Like GANs, GAIL is prone to training instability. The Adversarial Motion Prior (AMP) algorithm \cite{peng_amp_2021} is an extension of GAIL that incorporates several methods to improve stability in adversarial training. While the original AMP algorithm was used to generate physics-based stylized character animation, it was also used by \citet{sestini_automated_2022} for automated playtesting combining RL, IL, and curiosity driven learning. A notable example is \citet{ferguson2022imitating}, who propose the use of dynamic time warping IL to imitate distinct playstyles. This approach differs from ours in that it creates individual personas, rather than allowing for behaviors blending.

Similar to our work, Policy Fusion (PF) was proposed to combine distinct behavioral policies to obtain a meaningful ``fusion'' policy~\cite{fusion}. However, PF requires training one model for each persona and run them simultaneously during inference. In contrast, the MultiGAIL algorithm trains \textit{one model} in total, significantly reducing inference time. Furthermore, our algorithm is able to define different levels of blending, while PF has only one level. Finally, MultiGAIL works well with both discrete and continuous action spaces, while PF is limited to a discrete one.
\section{Method}
\label{sec:method}
Here we detail our approach to training agents which can switch and interpolate between several different personas.

\begin{figure}
    \centering
    \includegraphics[width=0.9\columnwidth]{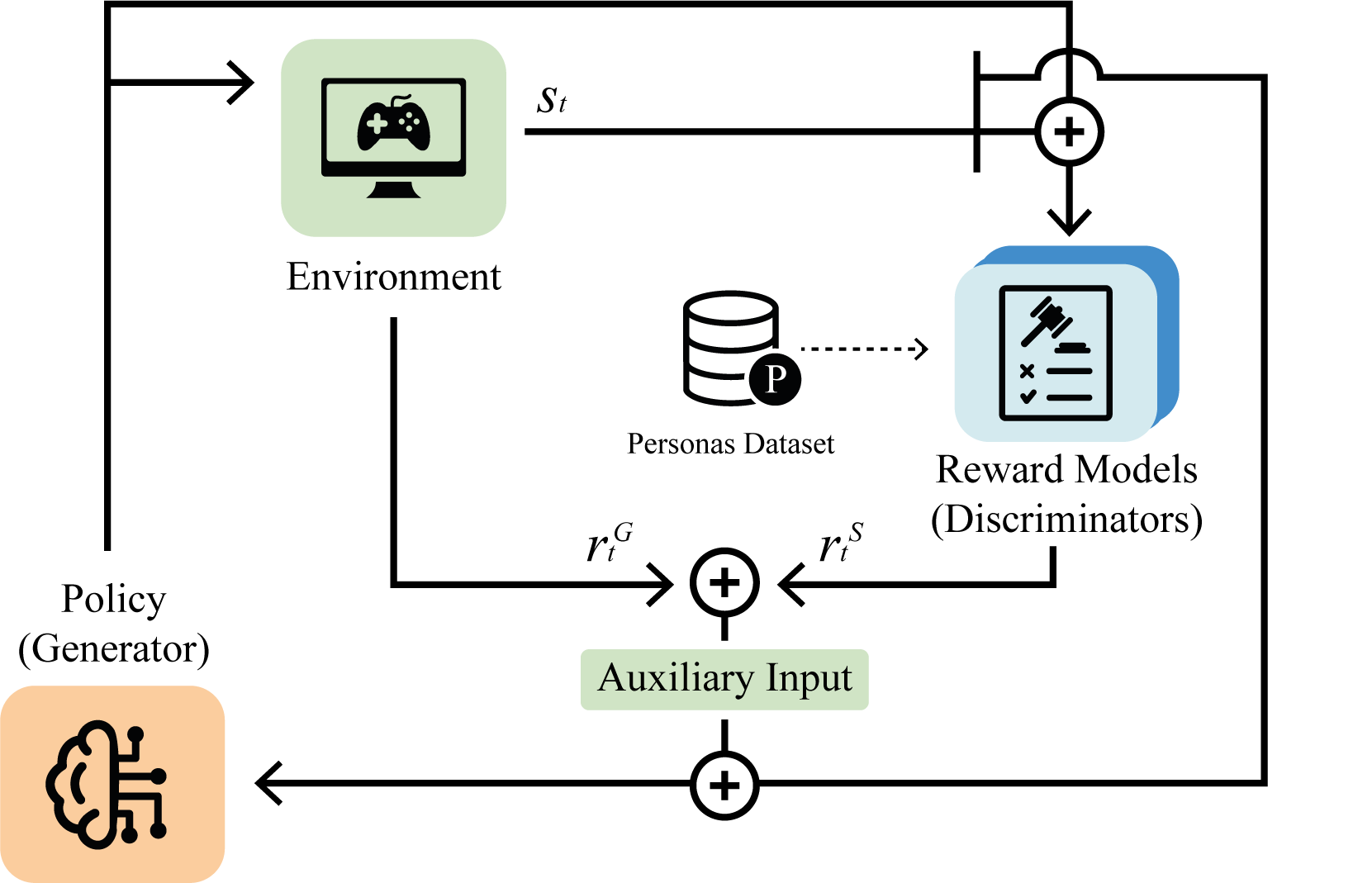}
    \caption{An overview of the MultiGAIL algorithm. Multiple discriminators learn to distinguish distinct personas generated by a single agent policy, while an auxiliary input enables the user to continuously sample a behavior from the behavior space induced by the demonstrations. Further details regarding the method are provided in Section~\ref{sec:method} and in Algorithm~\ref{alg:multigail}.}
    \label{fig:multi_gail}
\end{figure}

\subsection{Problem Statement}
Given a set of $n$ demonstrations $\mathbf{M} = \{M_i\}_{i=1}^{n}$ with each element $M_i$ representing a distinct persona, e.g. \textit{``act stealthy''} or \textit{``play aggressively''}, we want to train a \textit{single} policy $\pi$ capable of executing, switching and interpolating between the multiple personas. $M_i$ is a recording of demonstrations in the form: 
\begin{equation}
    \label{eq:playstyles}
    M_i = (s_1^i, a_1^i, s_2^i, a_2^i, ... , s_T^{i}),
\end{equation}
where $s_t^i$ is a state and $a_t^i$ is an action performed by the expert in the particular style $M_i$. To learn a policy which embody a persona involves not only specifying \textit{what} the agent should accomplish, but also \textit{how} they should achieve it \cite{microsoftblog}. Requiring game designers to adjust a reward function which incorporates both goal and style without any RL expertise is unrealistic. For non-experts in the RL field, providing demonstrations of a particular persona may be easier than defining a reward function representing different behaviors. A key requirement of MultiGAIL is the ease-of-use. The model should allow the user to intuitively access different distinct playstyles, or create new ones by sampling from the continuum between spanned each demonstration dataset. All at inference time. In the following sections, we describe the two main components of MultiGAIL: reward models learning and policy learning.

\begin{algorithm}[t]
\scriptsize
\caption{MultiGAIL training}\label{alg:multigail}
\begin{algorithmic}[0]
\Require $\mathbf{M} = \{M_i\}_{i=1}^{n}$ $(n$ persona demonstrations)
\State $\pi \gets$ initialize policy
\State $V \gets$ initialize value function
\State $\{D_i\}_{i=1}^{n} \gets$ initialize $n$ discriminators
\State $\alpha \gets$ initialize auxiliary inputs set 
\State $\mathcal{B} \gets$ initialize replay buffer
\While{not done}
\For{trajectory $i=1, ...,m$}
    
    \State $\tau^i \gets \{(s_t,a_t,r_t^G)_{t=1}^{T}, s_T^G, g\} $
    \State Sample $\{\alpha_i\}_{i=1}^{n}$ from  $\alpha$
    \For{time step $t = 1, ...,T$}
        \State $\{r_i^t\}_{i=1}^{n} = \{D_i(s_t,a_t)\}_{i=1}^{n}$
        \State $r_t^S = \sum_{i=1}^{n} \alpha_i r_i^t$
        \State $r_t = w^G r_t^G + w^S r_t^S$
        \State store $r_t$ in $\tau_i$
    \EndFor
    \State store $\tau^i$ in  $\mathcal{B}$
\EndFor
\State update $\{D_i\}_{i=1}^{n}$ with $\mathcal{L}^{\text{AMP}}$ with samples from $\mathcal{B}$
\State update update $\pi$ with $\mathcal{L}^{\text{PPO}}$ with samples from $\mathcal{B}$
\EndWhile
\end{algorithmic}
\end{algorithm}

\subsection{Reward Models Learning}
Our approach is an extension of the AMP algorithm \cite{peng_amp_2021}. The agent reward is split into two components, the task-reward $r^G$ and the style-reward $r^S$. The task-reward specifies higher-level objectives of the environment, and encodes \emph{what} the agent should accomplish, e.g. reach the goal or defeat all the enemies. The style-reward represents task-agnostic low-level details of the agent's behavior, essentially \textit{how} an objective should be accomplished. The agent's persona, e.g. playing aggressively or passively are associated with the style-reward. The total agent reward is a weighted linear combination of these two terms:
\begin{equation}
    r(s_t,a_t,s_{t+1}) = w^G r^G(s_t,a_t,s_{t+1},g)+ w^S r^S(s_t,a_t),
    \label{eq:total_reward}
\end{equation}
where $s_t$ is the state at timestep $t$, $a_t$ is the action performed at state $s_t$, $g$ is the particular environment goal, and $w^G$ and $w^S$ are coefficients. In our experiments, we set $w^G=w^S=1$. While it is possible to design the task-reward by hand, the style-rewards is considerably more challenging to construct. Style is naturally described qualitatively with observable characteristics, but trying to translate it to a quantitative reward function can easily result in unexpected behaviors~\cite{amodei_concrete_2016}. MultiGAIL utilizes adversarial training to model the style-rewards, avoiding the need to explicitly construct them. A single style-reward is represented with a discriminator function $D$ which is trained to differentiate whether a given state-action pair came from the agent policy or the expert demonstration $M_i$. The discriminator returns a probability value indicating the similarity between the demonstration data and the given state-action pair. The agent policy is thought of as a generator function and learns to mimic the expert demonstrations to generate state-action pairs to fool the discriminator and receive higher rewards. 

Our main contribution to the AMP algorithm was to extend the style-reward model to incorporate a set of discriminator functions $\mathbf{D} = \{D_i\}_{i=1}^{n}$, one for each demonstration dataset $M_i$, allowing a single policy to handle several different playstyles. Each discriminator returns a probability value indicating if a state-action pair comes from the policy or the expert persona, and all the probabilities are added and scaled by an auxiliary input, producing the new style-reward:
\begin{equation}
    r^S(s_t,a_t) = \sum_{i=1}^{n} \alpha_i \max \left[ 0, 1-0.25(D_i(s_t,a_t)-1)^2 \right].
\end{equation}
The auxiliary input $\alpha_i$ plays an essential role in policy learning and will be further detailed in the subsequent section. The parameters $\alpha_i$ are used to scale a reward value of for playstyle $M_i$, e.g. if we have two playstyles $M_1$ and $M_2$ and with parameters $\alpha_1 = 0$ and $\alpha_2 = 1$ that means that the agent will get $0$ reward for following $M_1$ while $r^S$ reward for following $M_2$. Each discriminator $D_i$ is always unaware of the particular auxiliary input $\alpha_i$ used at each episode. 
During training, we feed each discriminator $D_i$ with the same batch of policy experience. Each single one is tasked to distinguish between this batch and their own assigned set of demonstrations and persona $M_i$, even if the policy was instructed to simulate a persona different from $M_i$ in that batch. Our results in Section~\ref{sec:results} show this solution increases the robustness of MultiGAIL. Each discriminator is trained to optimize the Least-Square GAN (LSGAN) \cite{Mao_2017_ICCV} loss, and makes use of a gradient penalty to improve training stability compared to the conventional GAN loss: 
\begin{equation}
\begin{split}
    \mathcal{L}_i^{\text{AMP}} = \underset{D_i}{\argmin}\, \mathbb{E}_{d^{M_i}(s,a)} \left[\left(D_i(s,a)-1\right)^2\right] \\
    +\mathbb{E}_{d^\pi(s,a)} \left[ \left(D_i(s,a)+1\right)^2 \right] \\+ 
    \frac{w^{\text{gp}}}{2} \mathbb{E}_{d^{M_i}(s,a)} \left[\norm{\gradient_\phi D_i(\phi) \rvert_{\phi = (\Phi(s),\Phi(a))}  }^2\right], 
\end{split}
\end{equation}
where $d^M_i(s,a)$ and $d^\pi(s,a)$ respectively denote the likelihood of observing a state-action pair from the expert datasets or agent policy. The last term is the gradient penalty, and is scaled by the hyperparameter $w^{\text{gp}}$.

\begin{figure*}
    \centering
    \begin{tabular}{ccc}
         \includegraphics[width=0.3\textwidth]{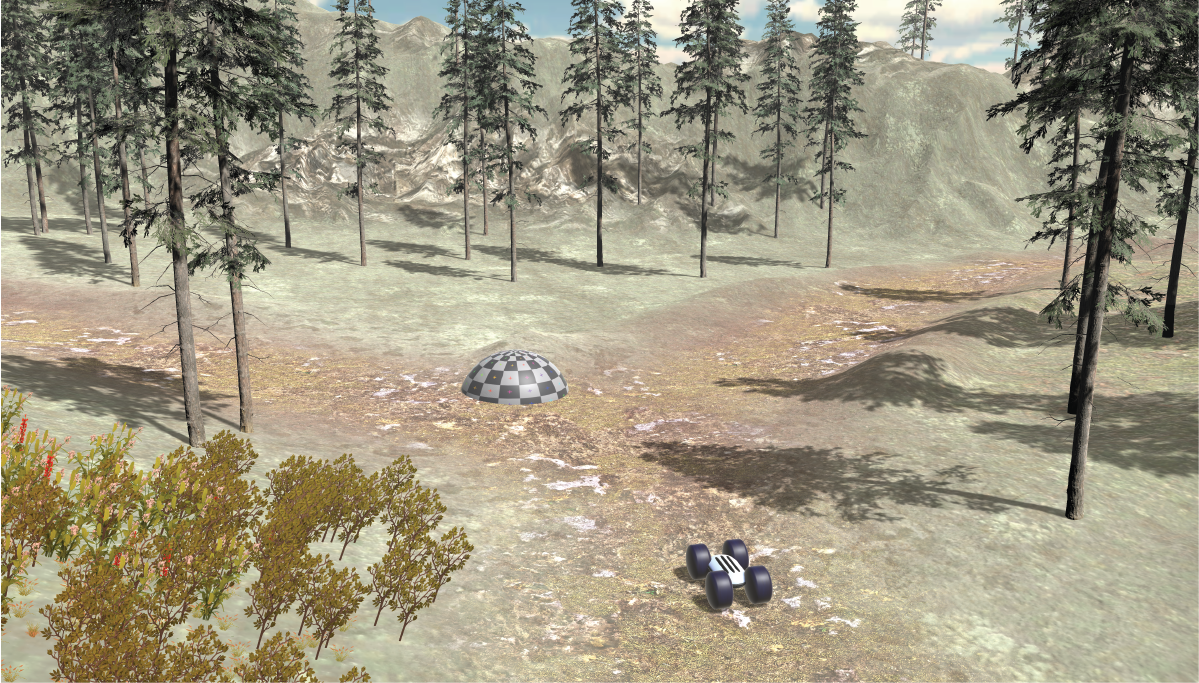} &
         \includegraphics[width=0.3\textwidth]{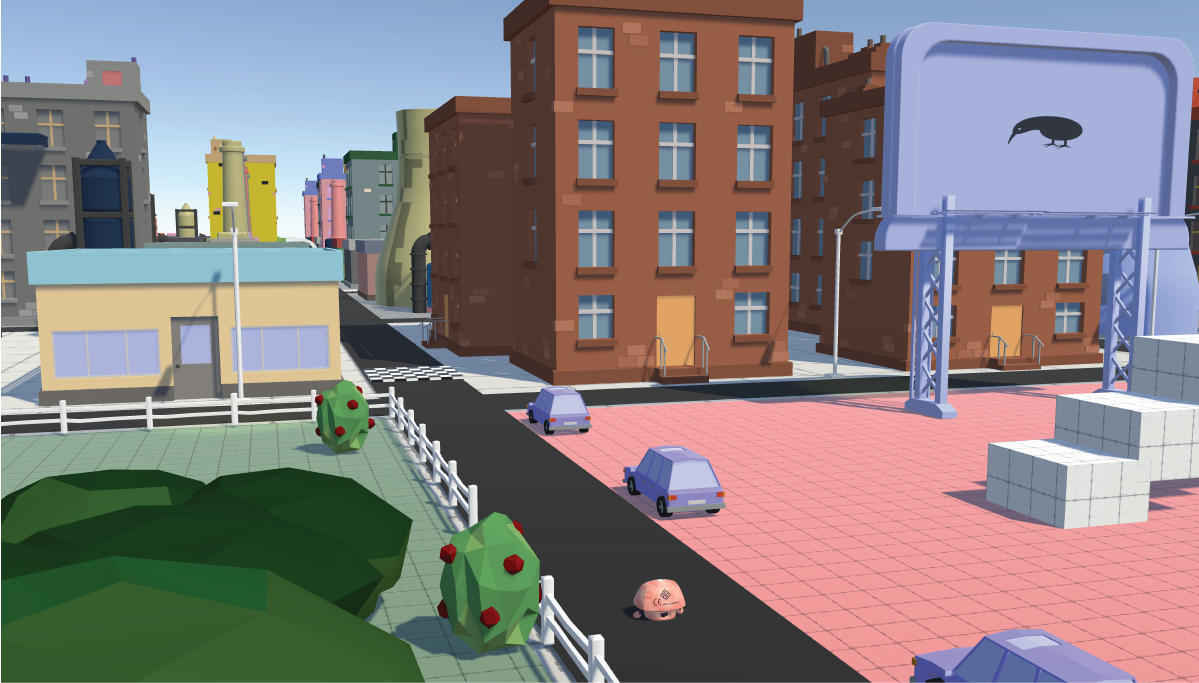} &
         \includegraphics[width=0.3\textwidth]{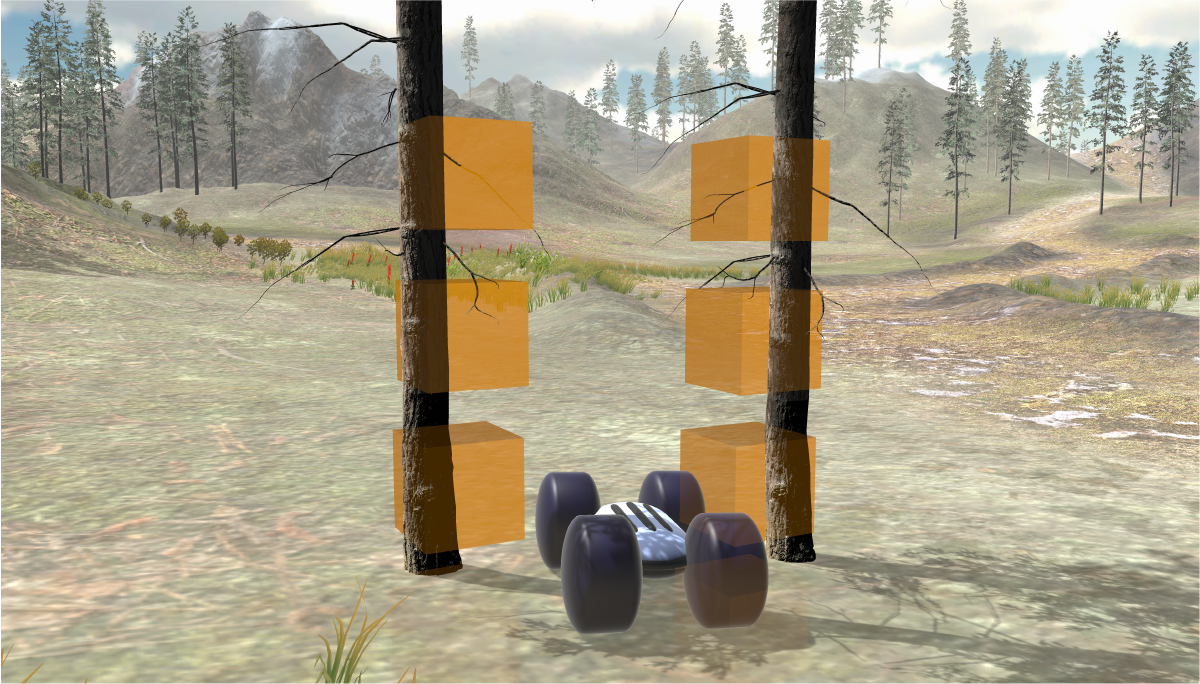}\\
         (a) & (b) & (c)
    \end{tabular}
    \caption{
    Screenshots of the environments used in this study. Both of them have the general objective of reaching a goal position. However, as noted in Section~\ref{sec:environments}, the focus of the paper is not the general objective, but rather \textit{how} the agent achieves it. (a) \textit{Racing Game}. This environment contains a two-dimensional continuous action space. Two different personas were used: \textit{careful} and \textit{reckless}. (b) \textit{Navigation Game}. In this environment the agent has a set of 9 discrete actions. Three different personas were utilized in the experiments: \textit{jump}, \textit{zigzag} and \textit{strafe}. (c) Example of semantic 3D map in the \textit{Racing Game}. Further information regarding the environments, personas and state spaces can be found in Section~\ref{sec:setup} and Section~\ref{sec:results}.
    }
    \label{fig:environments}
\end{figure*}

\subsection{Policy Learning}
Policy learning is done through Proximal Policy Optimization (PPO) \cite{schulman_proximal_2017} with the reward signal given by Equation~\eqref{eq:total_reward}. 
The overview of MultiGAIL training process is provided in Algorithm~\ref{alg:multigail}. It is similar to AMP~\cite{peng_amp_2021}, but with the addition of auxiliary control inputs and multiple discriminator models. The auxiliary input $\bar{\alpha} \in \mathbb{R}^n$ is a vector of values of the same size as the number of personas in $\mathbf{M}$, and each $\alpha_i$ are uniformly sampled from a predefined set of fractional values in $[0,1]$ before each training episode. The agent receives a task-reward from the environment at each time step, while a style-reward is given by the reward discriminator $D_i$ scaled by $\alpha_i$. As an example, suppose we have the case of two personas $(n=2)$, and a sampled value of $\bar{\alpha} = (\alpha_1 = 1,\; \alpha_2 = 0$) would indicate that the agent should fully mimic the persona in $M_1$ while disregarding the other personas datasets. A blended value, e.g. $\alpha_1 = 0.5,\; \alpha_2 = 1$ would steer the agent to incorporate characteristics found in both datasets, but value those from $M_2$ more. The rewards are combined according to Equation~\eqref{eq:total_reward}.

The main advantage of MultiGAIL is during inference. Once the model is trained, during testing game developers can select any desired $\bar{\alpha}$ values to control the behavior  of the personas, using only one single model. Furthermore, they can sample a behavior from a continuous behavior space defined by the set $\mathbf{M}$, increasing the ease-of-use of the model. This allows for the designers to utilize a variety of different personas by changing only one parameter at inference.

\section{Experimental Setup}
\label{sec:setup}
Here we detail the main components used in the evaluation of MultiGAIL. This includes two environments, the neural network architecture, and the training setup.

\subsection{Environments}
\label{sec:environments}
We used two environments to train and validate the MultiGAIL approach: the \textit{Racing Game} and the \textit{Navigation Game}. They are 3D game environments with the former having a continuous action space, while the latter has a discrete one. Further details regarding the action spaces are discussed in the following paragraphs. They both have the same overall environment objective which is to reach a goal area. However, the primary focus of our experiments is not to teach the agent to realize the environment goal, but rather observe how well agents adhere to the playstyle represented in the demonstrations.

The state space used by the agents of the two environments is conceptually the same, and was proposed by \citet{sestini_towards_nodate}. To allow the state space to facilitate the agent learning to mimic demonstrations, we first define a game goal position. Relative spatial information between the goal and agent position are represented as the $\mathbb{R}^2$ projections of the agent-to-goal vector onto the $XY$ and $XZ$ planes. The vector lengths are normalized by the gameplay area size. Additionally, the user also specifies a list of entities with objects of interest for the agent, including enemies, moving objects and intermediate goals. The objects in the entity list are represented in the same way as the main goal position, using relative spatial information. The state space also consists of game specific state observations, such as whether the agent is in the air, or its velocity. Game specific states for each environment are thoroughly stated in the following paragraphs. For local perception a 3D semantic map is used. An example of a semantic 3D occupancy map is illustrated in Figure~\ref{fig:environments}(c) as implemented in the \textit{Racing Game}. The map discretizes the surrounding space around the agent with voxels. The element which occupies a voxel's space is categorically described as an integer value to discern which object it is. For both environments, we use a $5 \times 5 \times 5$ centered in the position of the agent.

\minisection{Racing Game.}\label{sec:racing_game}
The \textit{Racing Game} takes place in a forested race track, and can be seen in Figure \ref{fig:environments}(a). The agent can concurrently control both acceleration and steering, thus having two degrees-of-freedom. Action values are normalized and bounded between $[-1,1]$, and are continuous within those limits. The game specific states include the velocity magnitude and its $xyz$ components, and the angular velocity in the driving plane. For this environment, we train two distinct personas:
\begin{itemize}
    \item \textit{careful}: applies forward acceleration sparingly, and uses minimal right and left steering. Moving in a smooth trajectory towards the goal;
    \item \textit{reckless}: maximum forward acceleration with excessive steering right and left. Moving in a zigzag pattern towards the goal. 
\end{itemize}

\minisection{Navigation Game.}\label{sec:navigation_game}
The \textit{Navigation Game} takes place in a open world city, and can be seen in Figure \ref{fig:environments}(b). The agent has a discrete action space of size 9 consisting of: move forward, move backwards, rotate right, rotate left, jump, shoot, sidestep right, sidestep left and do nothing. The game specific states include information of if the agent is climbing, has contact with the ground, is in an elevator, the jump cool down, and weapon magazine status. For this environment, we learn three distinct personas:
\begin{itemize}
    \item \textit{jump}: uses a combination of the jump and forward action to reach the goal;
    \item \textit{zigzag}: alternates turning agent left and right when moving, making the path look like a zigzag pattern; and
    \item \textit{strafe}: sidesteps left and right when moving forward.
\end{itemize}

\begin{figure}
    \centering
    \includegraphics[width=0.9\columnwidth]{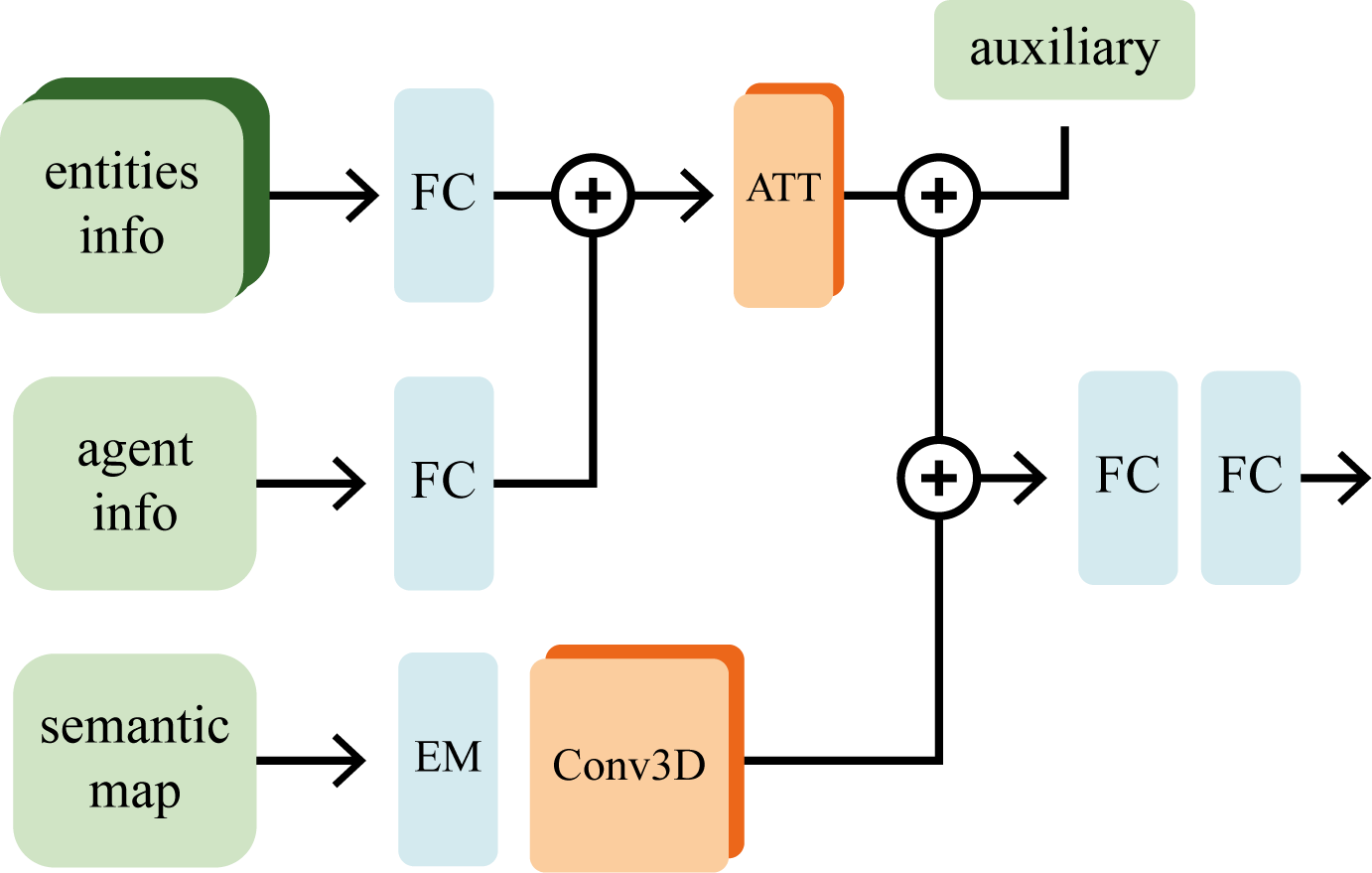}
    \caption{A schematic overview of the neural network used for the policy and the discriminators. The network parameters are described in detail in Section~\ref{sec:net}.}
    \label{fig:net}
\end{figure}

\subsection{Neural Network} \label{sec:net}
As we detailed in Section~\ref{sec:method}, our agent consists of two main components: the policy $\pi$ and a set of discriminators $\mathbf{D} = \{D_i\}_{i=1}^{n}$. The same network architecture is used for both components. It is inspired by the work of \citet{sestini_towards_nodate} and the complete neural network is shown in Figure~\ref{fig:net}. First, agent and goal information is passed into a linear layer ReLU activation, producing the self-embedding $x_\text{a} \in \mathbb{R}^{d}$, where $d$ is the embedding size. For our experiments, we set $d=128$. The list of entities is passed through a separate linear layer with shared weights producing the embedding $x_{\text{e}_i} \in \mathbb{R}^{d}$, one for each entity $\text{e}_i$ in the list. Each of these embedding vectors is concatenated with the self-embedding, producing $x_{\text{ae}_i} = [x_\text{a}, x_{\text{e}_i}]$, with $x_{\text{ae}_i} \in \mathbb{R}^{2d}$. This list of vectors is then passed through a transformer encoder with $4$ heads and average pooled, producing the vector $x_\text{t} \in \mathbb{R}^{2d}$. In parallel, the semantic occupancy map $M \in \mathbb{R}^{5 \times 5 \times 5}$ is first fed into an embedding layer of size 8 with $\tanh$ activation, transforming categorical representations into continuous ones, and then into a 3D convolutional network with three convolutional layers with 32, 64 and 128 filters, stride 2 and leaky ReLU activation. The output of this convolutional network is a vector embedding $x_\text{M} \in \mathbb{R}^{d}$ that is finally concatenated with $x_\text{t}$ producing $x_\text{Mt}$. For the policy, we then pass $x_\text{Mt}$ through one linear layer of size 256 and ReLU activation, and one final layer producing the action probability distribution. For discriminators, we pass $x_\text{Mt}$ through a feed forward network with two linear layers of size 256 and leaky ReLU activation. The last layer is a linear layer with sigmoid activation representing the discriminator probability. 

\subsection{Training Setup}
We implemented MultiGAIL with Tensorflow. For each one of the two personas of the \textit{Racing Game} we recorded ${\approx}10000$ state-action samples of expert demonstrations, for a total of~${\approx}20000$. For each one of the three personas of the \textit{Navigation Game} we recorded ${\approx}20000$ state-action samples of expert demonstrations, for a total of ${\approx}60000$. Further details regarding the qualitative behavior of these personas are provided in Section~\ref{sec:results}. All samples were recorded by humans. For all the experiments, we trained the agents until discriminators convergence. All training was performed by deploying ten agents in parallel on a NVIDIA TITAN V GPU with 12GB VRAM, a AMD Ryzen Threadripper 1950X 16-Core CPU and 64GB of RAM.

\section{Evaluation and Results}
\label{sec:results}
As mentioned in Section~\ref{sec:setup}, we are primarily interested in how well MultiGAIL is able to generate new playstyles by interpolating the personas defined by expert demonstrations. We also want to validate the algorithm's ability to use a single model able to switch between several distinct playstyle. Experiments were conducted in the two environments detailed in Section~\ref{sec:environments}. In both environments the agent is not only tasked with reaching the goal, but must also emulate expert demonstration playstyles.
\begin{table}
\centering
\caption{Divergence metrics of probability action distributions between expert demonstrations and agents trained either with MultiGAIL or AMP~\cite{peng_amp_2021}. MultiGAIL is trained as a single model and the results are sampled with either auxiliary input $\bar{\alpha} = (1,0)$, \textit{careful}; or $\bar{\alpha} = (0,1)$, \textit{reckless}. The AMP models are trained separately. The down arrow means lower is better. Abbreviations: KL (Kullback-Leibler), JS (Jensen-Shannon), $\chi^2$ (Pearson's $\chi^2$), W (Wasserstein).}
\label{tab:divergence}
\scalebox{0.88}{
\begin{tabular}{c|cc|cc}
\toprule
\multicolumn{1}{c|}{\multirow{2}{*}{Metrics $\downarrow$}} & \multicolumn{2}{c|}{\textit{careful}} & \multicolumn{2}{c}{\textit{reckless}} \\ 
\multicolumn{1}{c|}{}                                  & \multicolumn{1}{l}{MultiGAIL} & \multicolumn{1}{l|}{AMP \cite{peng_amp_2021}} & \multicolumn{1}{l}{MultiGAIL} & \multicolumn{1}{l}{AMP \cite{peng_amp_2021}} \\ \cline{1-5}
KL                                         & \textbf{3.75}                                     & 19.44                             & \textbf{3.40}                                    & 4.07                              \\
JS                                          & \textbf{0.37}                                      & 0.50                              & \textbf{0.54}                                     & 0.58                              \\
$\chi^2$                                      & 0.00                                    & 0.00                              & \textbf{65.01}                                    & 130.91                            \\
W                                             & 0.28                                     & \textbf{0.26}                               & \textbf{0.14}                                     & 0.19                              \\ \bottomrule
\end{tabular}}
\end{table}
\subsection{Single Model Multitask Policy}
In this section, we investigate the ability of MultiGAIL to exactly imitate the underlying style found in the expert demonstrations. To do so, we compare our algorithm to a single model baseline policy using the AMP algorithm~\cite{peng_amp_2021}. We first train an agent with Algorithm~\ref{alg:multigail} and then we evaluate it by running 30 episodes in a testing environment and storing the action distributions. To evaluate the ability of the agent trained with MultiGAIL to replicate the expert personas, we use a one-hot auxiliary input. For example, if there are two personas, we run the algorithm with $\bar{\alpha} = (1, 0)$ and then $\bar{\alpha} = (0, 1)$. This allows us to ensure that the agent follows the personas stored in the expert demonstrations without blending their behaviors. For the baseline, we train one model for each persona and then run 30 test episodes with each of the models. The action distributions of each approach is compared to the expert demonstrations and is reported in terms of Kullback-Leibler divergence, Jensen-Shannon divergence, Pearson's $\chi^2$-test, and the Wasserstein divergence.

\minisection{Racing Game.} As discussed in Section~\ref{sec:environments}, for the \textit{Racing Game} we have two distinct playstyles: \textit{careful} and \textit{reckless}, hence we have a MultiGAIL agent with an auxiliary input of size $2$ and two baseline models. Table~\ref{tab:divergence} shows the comparison between the action distributions. As the table shows, we can see that MultiGAIL is able to more accurately replicate the action distribution of the expert compared to the baseline. This is an interesting finding: the agent trained with both persona demonstrations at the same time is able to better replicate the individual styles compared to a set of models specifically trained to replicate each single one. However, this is in line with other works that explored the advantage of using multiple discriminators in adversarial training, as it can prevent mode collapse, since the generator has to optimize for several discriminators, making the task harder \cite{xu_gan-like,choi_mclgan}. Our hypothesis is that the 1 vs. $N$ generator discriminator setup improves learning compared to having separate models.

\minisection{Navigation Game.} We repeat the same test as for the \textit{Racing game}. Table~\ref{tab:percentage} shows the results. We can see that both the single models and the MultiGAIL successfully reflected the personas in the demonstrations, indicating that our approach has the same performance as the single models, but only requires \textit{training and using only one model}.
\setlength{\tabcolsep}{3pt}
\begin{table}
    \centering  
    \caption{Divergence metrics of probability action distributions between expert demonstrations and agents trained either with MultiGAIL or AMP~\cite{peng_amp_2021}. MultiGAIL is trained as a single model and the results are sampled either with $\bar{\alpha} = (1,0, 0)$, \textit{jump}; $\bar{\alpha} = (0,1, 0)$, \textit{zigzag}; or $\bar{\alpha} = (0,0,1)$, \textit{strafe}. The AMP models are trained separately. The down arrow means lower is better. Abbreviations: KL (Kullback-Leibler), JS (Jensen-Shannon), $\chi^2$ (Pearson's $\chi^2$), W (Wasserstein).}
    \scalebox{0.88}{
    \begin{tabular}{c|cc|cc|cc}
         \toprule
         \multicolumn{1}{c|}{\multirow{2}{*}{Metrics $\downarrow$}} & \multicolumn{2}{c|}{\textit{jump}} & \multicolumn{2}{c|}{\textit{zigzag}} & \multicolumn{2}{c}{\textit{strafe}} \\ 
           & MultiGAIL & AMP \cite{peng_amp_2021} & MultiGAIL & AMP \cite{peng_amp_2021} & MultiGAIL & AMP \cite{peng_amp_2021}  \\
         \hline
         KL        & 0.15 & 0.14 & 0.38 & 0.38 & 0.16 & 0.15\\ 
         JS        & 0.04 & 0.04 & 0.11 & 0.11 & 0.44 & 0.43\\ 
         $\chi^2$  & 0.24 & 0.23 & 0.54 & 0.54 & 0.25 & 0.25\\ 
         W         & 0.10 & 0.09 & 0.15 & 0.15 & 0.10 & 0.10\\ 
         \bottomrule
    \end{tabular}
    }
    \label{tab:percentage}
\end{table}
\setlength{\tabcolsep}{5pt}
\begin{table}
    \centering
    \caption{The Pearson correlation between the elements of the auxiliary input parameter and their corresponding main persona: $\alpha_1, \text{rotate}$; $\alpha_2, \text{jump}$; $\alpha_3, \text{sidestep}$. We can see that the associated action is positively correlated with its auxiliary input.}
    \scalebox{0.88}{
    \begin{tabular}{l|ccr}
         \toprule
          & $\alpha_1$ & $\alpha_2$ & $\alpha_3$\\
         \hline
         Rotate & \textbf{0.64} & -0.45 & -0.38\\
         Jump & -0.13 & \textbf{0.65} & -0.53\\
         Sidestep & -0.23 & -0.53 & \textbf{0.62}\\
         \bottomrule
    \end{tabular}}
    \label{tab:correlation}
\end{table}

\begin{figure*}
    \centering
    \includegraphics[width=0.65\textwidth]{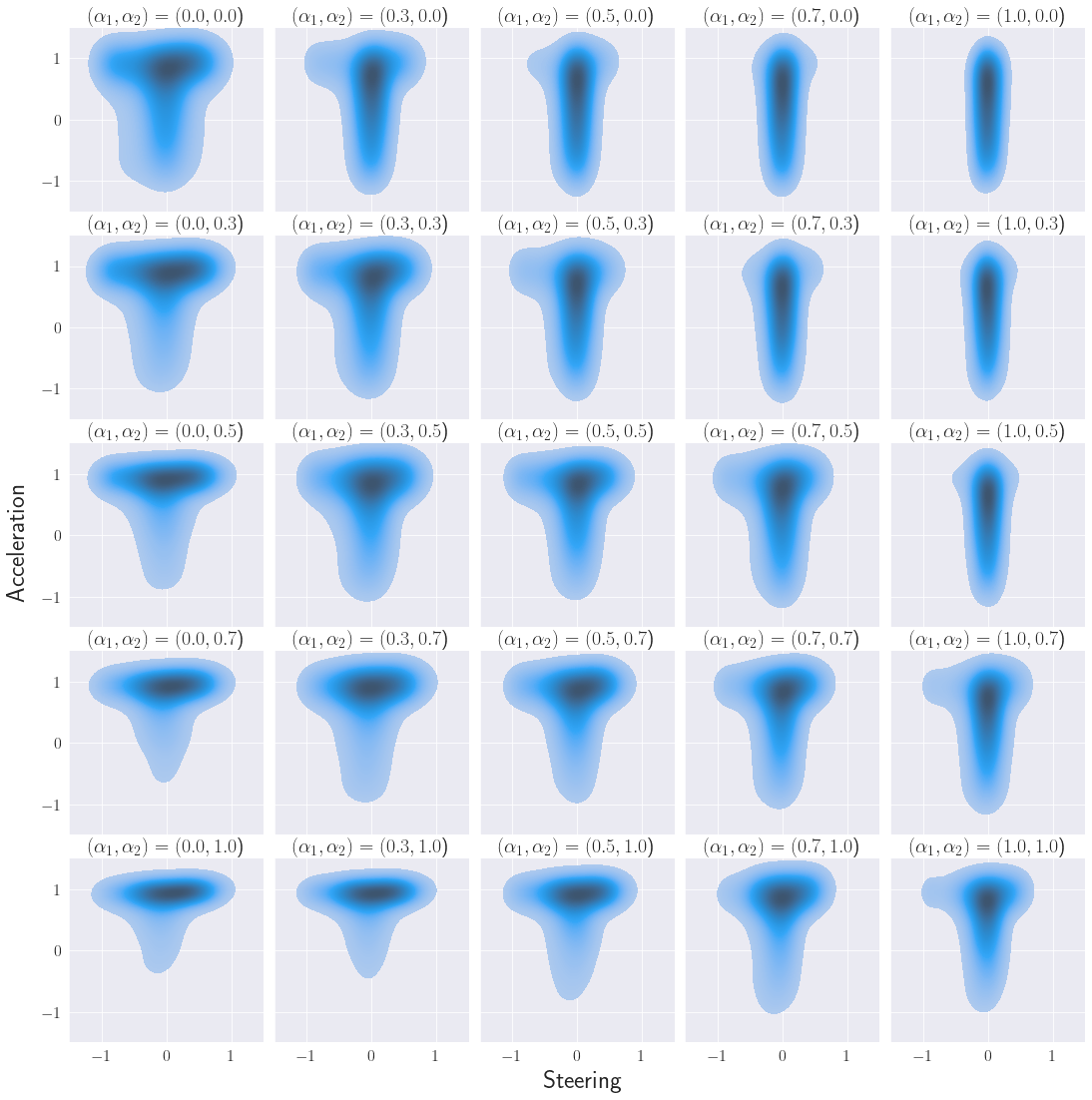}
    \caption{Kernel density estimation of the \textit{Racing Game} agent's sampled action distribution. Moving from left to right, we observe the steering action converging towards 0 while the acceleration action approaches -1. In contrast, traversing the plot from the top to the bottom, we can observe the acceleration action converging towards 1, while the steering action widens.}
    \label{fig:kde}
\end{figure*}

\subsection{Persona Interpolation}
For this evaluation, our primary interest is to be able to easily generate new personas by interpolating archetypal personas found in expert demonstrations. We train an agent with Algorithm~\ref{alg:multigail} and we evaluate it in a testing environment with different auxiliary input combinations, e.g. if there are two personas, we test it with a set of auxiliary inputs ranging from $(0, 0)$ to $(1, 1)$. For each combination, we test our agent for 30 episodes in a testing environment. As the baseline for this experiment we use the Policy Fusion (PF) technique~\cite{fusion}. However, as we mentioned in Section~\ref{sec:related}, PF is only usable in the discrete action space setting. Thus, we compare our method to this baseline only with the \textit{Navigation Game}.

\minisection{Racing Game.}
Qualitatively, we expect the \textit{careful} policy to avoid excessive steering, and to be conservative with full acceleration. In contrast, the \textit{reckless} playstyle steers excessively from side to side and heavily favors full acceleration. To understand how $\bar{\alpha}$ conditions the policy to draw from each playstyle, in Figure \ref{fig:kde} we plotted the joint action probability distribution for different auxiliary inputs. We can observe how the $\bar{\alpha}=(1,0)$ varies their acceleration, incorporating reverse acceleration, and using minimal steering, while $\bar{\alpha}=(0,1)$ policy favors acceleration and using a wide range of steering actions. The action probabilities agree with our descriptions of the \textit{careful} and \textit{reckless} driving styles. When the policy is conditioned with a $\bar{\alpha}$ with non-zero elements, features of both personas are mixed. Intuitively, increasing $\alpha_2$ from $[0, 1]$ incorporates more qualities of the \textit{careful} playstyle, decreasing wide steering, and conditions the agent to use smaller acceleration actions, and the opposite is also true. When observing the agent, we can see how it still steers like the \textit{reckless}, but with a more controlled acceleration. For the case when both elements are zero, our hypothesis was that the policy would be random as the agent does not receive any reward to shape its behavior; however, the action distribution looks like an average of the two singular playstyles. This is an unexpected result and we plan to study it further. It is worth mentioning that even if we change the auxiliary input values, the general performance of the agent does not decrease.
\begin{table*}
    \centering
    \caption{MultiGAIL vs. Policy Fusion~\cite{fusion}. The table shows the percentage of action usage relative to the personas made by the two approaches. We can see that MultiGAIL creates a better combination while being more robust.}
    \scalebox{0.88}{
    \begin{tabular}{cc|ccc|ccc}
         \toprule
          & & \multicolumn{3}{c|}{MultiGAIL} & \multicolumn{3}{c}{Policy Fusion~\cite{fusion}} \\
         \hline
         Styles & Auxiliary Style Input & Jump & Sidestep & Rotate  & Jump & Sidestep & Rotate \\
         \hline
         \textit{jump, strafe}     & $(1, 1, 0)$ & $0.04 \pm 0.03$ & $\mathbf{0.08 \pm 0.03}$ & $0.00 \pm 0.00$ & $\mathbf{0.26 \pm 0.02}$ & $0.00 \pm 0.00$ & $0.00 \pm 0.00$\\ 
         \textit{strafe, zigzag}     & $(0, 1, 1)$ & $0.00 \pm 0.00$ & $\mathbf{0.13 \pm 0.04}$ & $\mathbf{0.38 \pm 0.08}$ & $0.00 \pm 0.00$ & $0.09 \pm 0.03$ & $0.16 \pm 0.03$\\
         \textit{jump, zigzag}     & $(1, 0, 1)$ & $\mathbf{0.40 \pm 0.07}$ & $0.01 \pm 0.01$ & $\mathbf{0.21 \pm 0.10}$ & $0.18 \pm 0.04$ & $0.00 \pm 0.00$ & $0.11 \pm 0.03$\\
         \textit{jump, strafe, zigzag}    & $(1, 1, 1)$ & $0.05 \pm 0.03$ & $\mathbf{0.13 \pm 0.04}$ & $\mathbf{0.24 \pm 0.09}$ & $\mathbf{0.13 \pm 0.03}$ & $0.03 \pm 0.03$ & $0.08 \pm 0.01$\\
         \bottomrule
    \end{tabular}
    }
    \label{tab:multi_vs_pf}
\end{table*}

\minisection{Navigation Game.}
The demonstrated personas are characterized by how they favor certain actions. The \textit{jump} persona uses predominately the jump action, while the \textit{zigzag} and \textit{strafe} favor moving side to side by either rotating or sidestepping right and left, respectively. Evaluating each persona is done by running the agent for 30 episodes in a testing environment with different combinations of auxiliary inputs. Table~\ref{tab:correlation} illustrates the correlation between elements of the auxiliary input and their corresponding main action. We observe that increasing one auxiliary input strengthens the behavior of the associated persona while weakening the other ones. For instance, increasing $\alpha_1$ strengthens the \textit{jump} behavior while weakening the \textit{zigzag} and \textit{strafe} personas. This evaluation suggests that MultiGAIL is successfully able to represent the actions associated with the desired style. An intriguing aspect that we identified through empirical analysis is that the correlation between personas and auxiliary input is not linear, as opposed to the \textit{Racing Game}. For instance, an auxiliary input in the form $\bar{\alpha} = (0.3, 0.3, 0.3)$ does not result in a perfect blending, with certain personas being stronger than others. We intend to further investigate this finding in the future.

To further evaluate our approach in the discrete setting, we compare MultiGAIL with PF~\cite{fusion}. We aim to understand which method is better at representing the combination of different personas. Since PF is unable to define different levels of blending, we compare it with MultiGAIL using extreme auxiliary inputs: for instance, an auxiliary input of $\bar{\alpha} = (1, 1, 1)$ for blending all three personas at the same level. Table~\ref{tab:multi_vs_pf} shows the percentage of action usage relative to the personas made by the two approaches. We observe that the MultiGAIL agent can better transition between different personas on average.

\section{Conclusion and Limitations}
In this paper, we introduced Multimodal Generative Adversarial Imitation Learning (MultiGAIL), a novel imitation learning algorithm that trains a single policy to imitate diverse behavioral styles, or so-called personas. MultiGAIL exploits a combination of an auxiliary input and multiple discriminators to concurrently learn distinct qualitative behaviors that can be sampled at inference time by changing the auxiliary input to the agent. Furthermore, MultiGAIL enables the user to blend different personas together, interpolating new behaviors that were not present in the demonstrations. Our experimental results and analyses show that MultiGAIL is both capable of reproducing the underlying styles found in the given demonstrations as well as persona interpolation \textit{using only a single model}, for both continuous- and discrete-action space. We believe MultiGAIL offers a new avenue in training simple and easy-to-use agents that represent different playstyle, reducing the need for creating engineered reward functions and multiple models. Although the aim of MultiGAIL is to aid game designers to create game testing agents that better reflect the variety of actual players, we believe this approach would be beneficial for anyone who desires to create easy-to-use agents with distinct behavioral styles.

Despite the successes of MultiGAIL, there are limitations to consider. To validate our approach, we tested MultiGAIL on simple navigation tasks. In the future, we plan to use MultiGAIL in more complex environments. Our experimental analyses have revealed that the relation between auxiliary input and personas is not linear, particularly when we have more than two styles: for instance, $\bar{\alpha} = (0.5, 0.5)$ does not necessarily guarantee a perfect half combination. We find this relationship intriguing and we intend to study it further. Finally, future work includes training on more than three personas; it would be interesting to understand how well it scales according to the number of personas.

{\footnotesize \bibliography{refs_long}}

\begin{thebibliography}{31}
\providecommand{\natexlab}[1]{#1}
\providecommand{\url}[1]{#1}
\csname url@samestyle\endcsname
\providecommand{\newblock}{\relax}
\providecommand{\bibinfo}[2]{#2}
\providecommand{\BIBentrySTDinterwordspacing}{\spaceskip=0pt\relax}
\providecommand{\BIBentryALTinterwordstretchfactor}{4}
\providecommand{\BIBentryALTinterwordspacing}{\spaceskip=\fontdimen2\font plus
\BIBentryALTinterwordstretchfactor\fontdimen3\font minus
  \fontdimen4\font\relax}
\providecommand{\BIBforeignlanguage}[2]{{%
\expandafter\ifx\csname l@#1\endcsname\relax
\typeout{** WARNING: IEEEtranN.bst: No hyphenation pattern has been}%
\typeout{** loaded for the language `#1'. Using the pattern for}%
\typeout{** the default language instead.}%
\else
\language=\csname l@#1\endcsname
\fi
#2}}
\providecommand{\BIBdecl}{\relax}
\BIBdecl

\bibitem[Mnih et~al.(2013)Mnih, Kavukcuoglu, Silver, Graves, Antonoglou,
  Wierstra, and Riedmiller]{mnih_playing_2013}
V.~Mnih, K.~Kavukcuoglu, D.~Silver, A.~Graves, I.~Antonoglou, D.~Wierstra, and
  M.~Riedmiller, ``Playing atari with deep reinforcement learning,''
  \emph{arXiv preprint arXiv:1312.5602}, 2013.

\bibitem[Vinyals et~al.(2019)Vinyals, Babuschkin, Czarnecki, Mathieu, Dudzik,
  Chung, Choi, Powell, Ewalds, Georgiev, et~al.]{vinyals_grandmaster_2019}
O.~Vinyals, I.~Babuschkin, W.~M. Czarnecki, M.~Mathieu, A.~Dudzik, J.~Chung,
  D.~H. Choi, R.~Powell, T.~Ewalds, P.~Georgiev \emph{et~al.}, ``Grandmaster
  level in starcraft ii using multi-agent reinforcement learning,''
  \emph{Nature}, vol. 575, no. 7782, pp. 350--354, 2019.

\bibitem[Berner et~al.(2019)Berner, Brockman, Chan, Cheung, Debiak, Dennison,
  Farhi, Fischer, Hashme, Hesse, et~al.]{openai_dota_2019}
C.~Berner, G.~Brockman, B.~Chan, V.~Cheung, P.~Debiak, C.~Dennison, D.~Farhi,
  Q.~Fischer, S.~Hashme, C.~Hesse \emph{et~al.}, ``Dota 2 with large scale deep
  reinforcement learning,'' \emph{arXiv preprint arXiv:1912.06680}, 2019.

\bibitem[Stahlke et~al.(2019)Stahlke, Nova, and Mirza-Babaei]{playfulness}
S.~Stahlke, A.~Nova, and P.~Mirza-Babaei, ``Artificial playfulness: A tool for
  automated agent-based playtesting,'' in \emph{CHI Conference on Human Factors
  in Computing Systems}, 2019.

\bibitem[Cho et~al.(2010)Cho, Sohn, Park, and Kang]{scenariobased}
C.-S. Cho, K.-M. Sohn, C.-J. Park, and J.-H. Kang, ``Online game testing using
  scenario-based control of massive virtual users,'' in \emph{International
  Conference on Advanced Communication Technology (ICACT)}, 2010.

\bibitem[Sestini et~al.(2022{\natexlab{a}})Sestini, Gissl{\'e}n, Bergdahl,
  Tollmar, and Bagdanov]{sestini_automated_2022}
A.~Sestini, L.~Gissl{\'e}n, J.~Bergdahl, K.~Tollmar, and A.~D. Bagdanov,
  ``Automated gameplay testing and validation with curiosity-conditioned
  proximal trajectories,'' \emph{IEEE Transactions on Games}, 2022.

\bibitem[Gordillo et~al.(2021)Gordillo, Bergdahl, Tollmar, and
  Gissl{\'e}n]{gordillo_improving_2021}
C.~Gordillo, J.~Bergdahl, K.~Tollmar, and L.~Gissl{\'e}n, ``Improving
  playtesting coverage via curiosity driven reinforcement learning agents,'' in
  \emph{Conference on Games (CoG)}, 2021.

\bibitem[Dulac-Arnold et~al.(2021)Dulac-Arnold, Levine, Mankowitz, Li,
  Paduraru, Gowal, and Hester]{inefficiency}
G.~Dulac-Arnold, N.~Levine, D.~J. Mankowitz, J.~Li, C.~Paduraru, S.~Gowal, and
  T.~Hester, ``Challenges of real-world reinforcement learning: definitions,
  benchmarks and analysis,'' \emph{Machine Learning}, vol. 110, no.~9, pp.
  2419--2468, 2021.

\bibitem[Politowski et~al.(2022)Politowski, Gu{\'e}h{\'e}neuc, and
  Petrillo]{towardstest}
C.~Politowski, Y.-G. Gu{\'e}h{\'e}neuc, and F.~Petrillo, ``Towards automated
  video game testing: Still a long way to go,'' \emph{arXiv preprint
  arXiv:2202.12777}, 2022.

\bibitem[Aytemiz et~al.(2021)Aytemiz, Jacob, and Devlin]{microsoftblog}
B.~Aytemiz, M.~Jacob, and S.~Devlin, ``Acting with style: Towards
  designer-centred reinforcement learning for the video games industry,'' in
  \emph{CHI Workshop on Reinforcement Learning for Humans, Computer, and
  Interaction (RL4HCI)}, 2021.

\bibitem[Sestini et~al.(2022{\natexlab{b}})Sestini, Bergdahl, Tollmar,
  Bagdanov, and Gissl{\'e}n]{sestini_towards_nodate}
A.~Sestini, J.~Bergdahl, K.~Tollmar, A.~D. Bagdanov, and L.~Gissl{\'e}n,
  ``Towards informed design and validation assistance in computer games using
  imitation learning,'' in \emph{NeurIPS workshop on Human In the Loop
  Learning}, 2022.

\bibitem[Silver et~al.(2017)Silver, Hubert, Schrittwieser, Antonoglou, Lai,
  Guez, Lanctot, Sifre, Kumaran, Graepel, et~al.]{silver_mastering_2017}
D.~Silver, T.~Hubert, J.~Schrittwieser, I.~Antonoglou, M.~Lai, A.~Guez,
  M.~Lanctot, L.~Sifre, D.~Kumaran, T.~Graepel \emph{et~al.}, ``Mastering chess
  and shogi by self-play with a general reinforcement learning algorithm,''
  \emph{arXiv preprint arXiv:1712.01815}, 2017.

\bibitem[Mugrai et~al.(2019)Mugrai, Silva, Holmg{\aa}rd, and
  Togelius]{mugrai_automated_2019}
L.~Mugrai, F.~Silva, C.~Holmg{\aa}rd, and J.~Togelius, ``Automated playtesting
  of matching tile games,'' in \emph{Conference on Games (CoG)}, 2019.

\bibitem[Gudmundsson et~al.(2018)Gudmundsson, Eisen, Poromaa, Nodet, Purmonen,
  Kozakowski, Meurling, and Cao]{gudmundsson_human-like_2018}
S.~F. Gudmundsson, P.~Eisen, E.~Poromaa, A.~Nodet, S.~Purmonen, B.~Kozakowski,
  R.~Meurling, and L.~Cao, ``Human-like playtesting with deep learning,'' in
  \emph{Conference on Computational Intelligence and Games (CIG)}, 2018.

\bibitem[Bergdahl et~al.(2020)Bergdahl, Gordillo, Tollmar, and
  Gissl{\'e}n]{bergdahl_augmenting_2020}
J.~Bergdahl, C.~Gordillo, K.~Tollmar, and L.~Gissl{\'e}n, ``Augmenting
  automated game testing with deep reinforcement learning,'' in
  \emph{Conference on Games (CoG)}, 2020.

\bibitem[Zheng et~al.(2019)Zheng, Xie, Su, Ma, Hao, Meng, Liu, Shen, Chen, and
  Fan]{zheng_wuji_2019}
Y.~Zheng, X.~Xie, T.~Su, L.~Ma, J.~Hao, Z.~Meng, Y.~Liu, R.~Shen, Y.~Chen, and
  C.~Fan, ``Wuji: Automatic online combat game testing using evolutionary deep
  reinforcement learning,'' in \emph{International Conference on Automated
  Software Engineering (ASE)}, 2019, pp. 772--784.

\bibitem[Holmg{\aa}rd et~al.(2018)Holmg{\aa}rd, Green, Liapis, and
  Togelius]{holmgard_automated_2018}
C.~Holmg{\aa}rd, M.~C. Green, A.~Liapis, and J.~Togelius, ``Automated
  playtesting with procedural personas through mcts with evolved heuristics,''
  \emph{IEEE Transactions on Games}, vol.~11, no.~4, pp. 352--362, 2018.

\bibitem[Le~Pelletier~de Woillemont et~al.(2021)Le~Pelletier~de Woillemont,
  Labory, and Corruble]{de_woillemont_configurable_2021}
P.~Le~Pelletier~de Woillemont, R.~Labory, and V.~Corruble, ``Configurable agent
  with reward as input: a play-style continuum generation,'' in
  \emph{Conference on Games (CoG)}, 2021.

\bibitem[Amodei et~al.(2016)Amodei, Olah, Steinhardt, Christiano, Schulman, and
  Man{\'e}]{amodei_concrete_2016}
D.~Amodei, C.~Olah, J.~Steinhardt, P.~Christiano, J.~Schulman, and D.~Man{\'e},
  ``Concrete problems in ai safety,'' \emph{arXiv preprint arXiv:1606.06565},
  2016.

\bibitem[Bain and Sammut(1995)]{bc1}
M.~Bain and C.~Sammut, ``A framework for behavioural cloning.'' in
  \emph{Machine Intelligence 15}, 1995, pp. 103--129.

\bibitem[Ross et~al.(2011)Ross, Gordon, and Bagnell]{ross_reduction_2011}
S.~Ross, G.~Gordon, and D.~Bagnell, ``A reduction of imitation learning and
  structured prediction to no-regret online learning,'' in \emph{in
  international conference on artificial intelligence and statistics}, 2011.

\bibitem[Ho and Ermon(2016)]{ho_generative_2016}
J.~Ho and S.~Ermon, ``Generative adversarial imitation learning,''
  \emph{Advances in neural information processing systems}, vol.~29, 2016.

\bibitem[Goodfellow et~al.(2020)Goodfellow, Pouget-Abadie, Mirza, Xu,
  Warde-Farley, Ozair, Courville, and Bengio]{goodfellow_generative_2014}
I.~Goodfellow, J.~Pouget-Abadie, M.~Mirza, B.~Xu, D.~Warde-Farley, S.~Ozair,
  A.~Courville, and Y.~Bengio, ``Generative adversarial networks,''
  \emph{Communications of the ACM}, vol.~63, no.~11, pp. 139--144, 2020.

\bibitem[Finn et~al.(2016)Finn, Christiano, Abbeel, and
  Levine]{finn_connection_2016}
C.~Finn, P.~Christiano, P.~Abbeel, and S.~Levine, ``A connection between
  generative adversarial networks, inverse reinforcement learning, and
  energy-based models,'' \emph{arXiv preprint arXiv:1611.03852}, 2016.

\bibitem[Peng et~al.(2021)Peng, Ma, Abbeel, Levine, and
  Kanazawa]{peng_amp_2021}
X.~B. Peng, Z.~Ma, P.~Abbeel, S.~Levine, and A.~Kanazawa, ``{AMP}: adversarial
  motion priors for stylized physics-based character control,'' \emph{{ACM}
  Transactions on Graphics}, vol.~40, no.~4, pp. 1--20, 2021.

\bibitem[Ferguson et~al.(2022)Ferguson, Devlin, Kudenko, and
  Walker]{ferguson2022imitating}
M.~Ferguson, S.~Devlin, D.~Kudenko, and J.~A. Walker, ``Imitating playstyle
  with dynamic time warping imitation,'' in \emph{International Conference on
  the Foundations of Digital Games}, 2022.

\bibitem[Sestini et~al.(2021)Sestini, Kuhnle, and Bagdanov]{fusion}
A.~Sestini, A.~Kuhnle, and A.~D. Bagdanov, ``Policy fusion for adaptive and
  customizable reinforcement learning agents,'' in \emph{2021 Conference on
  Games (CoG)}.\hskip 1em plus 0.5em minus 0.4em\relax IEEE, 2021, pp. 01--08.

\bibitem[Mao et~al.(2017)]{Mao_2017_ICCV}
X.~Mao \emph{et~al.}, ``Least squares generative adversarial networks,'' in
  \emph{International Conference on Computer Vision (ICCV)}, 2017.

\bibitem[Schulman et~al.(2017)Schulman, Wolski, Dhariwal, Radford, and
  Klimov]{schulman_proximal_2017}
J.~Schulman, F.~Wolski, P.~Dhariwal, A.~Radford, and O.~Klimov, ``Proximal
  policy optimization algorithms,'' \emph{arXiv preprint arXiv:1707.06347},
  2017.

\bibitem[Xu and Karamouzas(2021)]{xu_gan-like}
P.~Xu and I.~Karamouzas, ``A gan-like approach for physics-based imitation
  learning and interactive character control,'' \emph{ACM on Computer Graphics
  and Interactive Techniques}, vol.~4, no.~3, pp. 1--22, 2021.

\bibitem[Choi and Han(2021)]{choi_mclgan}
J.~Choi and B.~Han, ``Mcl-gan: Generative adversarial networks with multiple
  specialized discriminators,'' \emph{arXiv preprint arXiv:2107.07260}, 2021.

\end{thebibliography}
\bibliographystyle{IEEEtranN}
\end{document}